\documentclass{article}
\usepackage{ijcai24}

\usepackage{times}
\usepackage{soul}
\usepackage{url}
\usepackage[hidelinks]{hyperref}
\usepackage[utf8]{inputenc}
\usepackage[small]{caption}
\usepackage{graphicx}
\usepackage{amsmath}
\usepackage{amsthm}
\usepackage{booktabs}
\usepackage{algorithm}
\usepackage{algorithmic}
\usepackage[switch]{lineno}
\usepackage{tikz-qtree}
\usepackage[edges]{forest}
\usepackage{pgf-pie}
\usepackage{enumitem}
\usepackage{xcolor}
\usepackage{bm}
\usepackage{multirow}
\usepackage{pgfplots}
\usepackage{amsmath, amssymb, mathtools}
\definecolor{customColor}{rgb}{0.933, 0.969, 1}
\definecolor{baseColor}{rgb}{0.886, 0.922, 0.980}
\definecolor{cpurple}{rgb}{0.675, 0.573, 0.922}
\definecolor{cblue}{rgb}{0.310, 0.757, 0.910}
\definecolor{cgreen}{rgb}{0.627, 0.835, 0.408}
\definecolor{corange}{rgb}{1, 0.808, 0.329}
\definecolor{cred}{rgb}{0.929, 0.333, 0.392}
\pgfplotsset{compat=1.18}

\linenumbers

\title{A Survey of Table Reasoning with Large Language Models}
\author{
    Xuanliang Zhang, Dingzirui Wang, Longxu Dou, Qingfu Zhu, Wanxiang Che
    \affiliations
    Research Center for Social Computing and Information Retrieval\\
    Harbin Institute of Technology, China
    \emails
    \{xuanliangzhang, dzrwang, lxdou, qfzhu, car\}@ir.hit.edu.cn
}

\begin{document}
\nolinenumbers

\maketitle

\begin{abstract}
    Table reasoning, which aims to generate the corresponding answer to the question following the user requirement according to the provided table, and optionally a text description of the table, effectively improving the efficiency of obtaining information.
    Recently, using Large Language Models (LLMs) has become the mainstream method for table reasoning, because it not only significantly reduces the annotation cost but also exceeds the performance of previous methods.
    However, existing research still lacks a summary of LLM-based table reasoning works.
    Due to the existing lack of research, questions about which techniques can improve table reasoning performance in the era of LLMs, why LLMs excel at table reasoning, and how to enhance table reasoning abilities in the future, remain largely unexplored. This gap significantly limits progress in research.
    To answer the above questions and advance table reasoning research with LLMs, we present this survey to analyze existing research, inspiring future work.
    In this paper, we analyze the mainstream techniques used to improve table reasoning performance in the LLM era\footnote{We summarize the detailed resources of the current research in \url{https://github.com/zhxlia/Awesome-TableReasoning-LLM-Survey}.}, and the advantages of LLMs compared to pre-LLMs for solving table reasoning.
    We provide research directions from both the improvement of existing methods and the expansion of practical applications to inspire future research.
\end{abstract}

\section{Introduction}
        \begin{figure}[ht]
    \centering
    \includegraphics[width=1.0\linewidth]{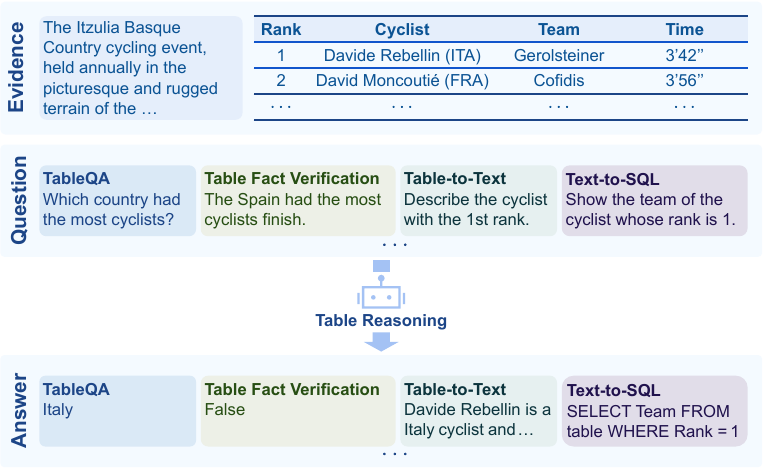}
    \caption{The illustration of various table reasoning tasks.}
    \label{fig:intro}
\end{figure}

Table reasoning task, which significantly improves the efficiency of obtaining and processing data from massive amounts of tables, plays an important role in the study of Natural Language Processing (NLP)~\cite{jin2022survey_tableqa}.
An illustration of the table reasoning task is shown in Figure~\ref{fig:intro}.
Given one or more tables, this task requires the model to generate results corresponding to the given question, as required by users (e.g., table QA~\cite{pasupat-liang-2015-compositional_wikitablequestions}, table fact verification~\cite{2019TabFactA}).

In the past, research in table reasoning has gone through several phases: rule-based, neural network-based, and pre-trained language model-based~\cite{jin2022survey_tableqa}, which we call the \textbf{pre-LLM era}.
Recent research~\cite{LLMSurvey} has shown that Large Language Models (LLMs) exhibit compelling performance across NLP tasks, in particular, dramatically reducing annotation requirements, which we call the \textbf{LLM era}.
Consequently, there has been a lot of works on applying LLMs to the table reasoning task to reduce overhead and outperform the methods in the pre-LLM era, which has become the current mainstream method.

\renewcommand{\dblfloatpagefraction}{.9}
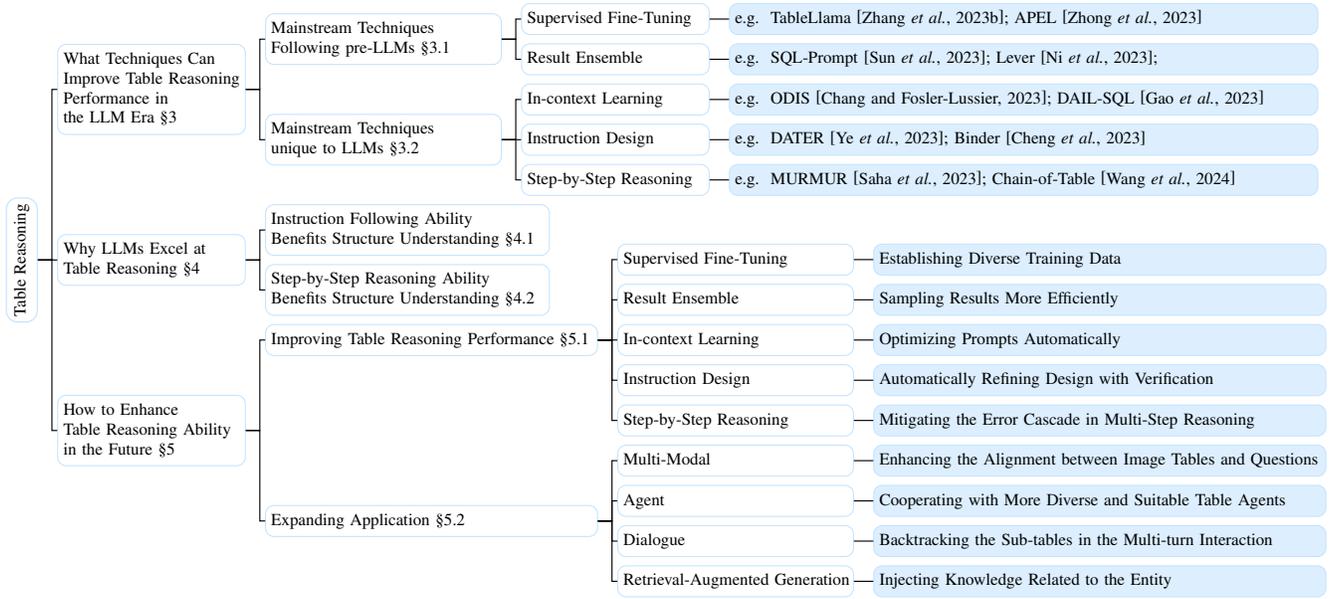
\begin{figure*}
    \tiny
    \resizebox{\textwidth}{!}{
        \tikzset{
    question/.style = {align=left,text width=2.2cm,rounded corners=3pt, fill=white,draw=customColor!360,line width=0.1mm, minimum height=0.3cm, anchor=center},  
    middle/.style = {align=left,text width=3.4cm,rounded corners=3pt, fill=white,draw=customColor!360,line width=0.1mm, minimum height=0.3cm, anchor=center}, 
    technique_middle/.style = {align=left,text width=2.8cm,rounded corners=3pt, fill=white,draw=customColor!360,line width=0.1mm, minimum height=0.3cm, anchor=center}, 
    future_middle/.style = {align=left,text width=4.0cm,rounded corners=3pt, fill=white,draw=customColor!360,line width=0.1mm, minimum height=0.3cm, anchor=center}, 
    technique/.style = {align=left,text width=2.2cm,rounded corners=3pt, fill=white,draw=customColor!360,line width=0.1mm, minimum height=0.3cm, anchor=center}, 
    future_leaf/.style = {align=left,text width=5.5cm,rounded corners=3pt, fill=customColor!200,draw=customColor!360,line width=0.1mm, minimum height=0.3cm, anchor=center},
    leaf/.style = {align=left, text width=7.2cm,rounded corners=3pt, fill=customColor!200,draw=customColor!360,line width=0.1mm, minimum height=0.3cm, anchor=center},  
}

\begin{forest}
    for tree={
        forked edges,
        grow=east,
        reversed=true,
        anchor=base west,
        parent anchor=east,
        child anchor=west,
        base=left,
        draw,
        rounded corners,
        s sep=3pt,
        align=center,
        parent/.style={rotate=90, child anchor=north, parent anchor=south, anchor=center, draw=customColor!360}
    },
    [Table Reasoning, parent
        [{\begin{tabular}[t]{@{}l@{}} What Techniques Can\\Improve Table Reasoning\\ Performance in\\ the LLM Era \S\ref{sec:methodology}\end{tabular}}, question
            [{\begin{tabular}[t]{@{}l@{}} Mainstream Techniques\\Following pre-LLMs \S\ref{subsec:Mainstream Techniques Following pre-LLMs}\\ \end{tabular}} , technique_middle
                [Supervised Fine-Tuning, technique
                    [e.g.\, TableLlama~\cite{zhang2023tablellama}; APEL~\cite{zhong-etal-2023-apel}, leaf]
                ]
                [Result Ensemble, technique
                    [e.g.\, {\begin{tabular}[t]{@{}l@{}}SQL-Prompt~\cite{sun-etal-2023-sqlprompt}; Lever~\cite{ni2023lever};\\\end{tabular}}, leaf]
                ]
            ]
            [{\begin{tabular}[t]{@{}l@{}} Mainstream Techniques\\ unique to LLMs \S\ref{subsec:Mainstream Techniques Unique to LLMs} \\ \end{tabular}} , technique_middle
                [In-context Learning, technique
                    [e.g.\, ODIS~\cite{chang-fosler-lussier-2023-selective_odis}; DAIL-SQL~\cite{gao2023dail}, leaf]
                ]
                [Instruction Design, technique
                    [e.g.\, DATER~\cite{dater}; Binder~\cite{binder}, leaf]
                ]
                [Step-by-Step Reasoning, technique
                    [e.g.\, MURMUR~\cite{saha-etal-2023-murmur}; Chain-of-Table~\cite{anonymous2023chainoftable}, leaf]
                ]
            ]
        ]
        [{\begin{tabular}[t]{@{}l@{}} Why LLMs Excel at\\Table Reasoning \S\ref{sec:discussion}\end{tabular}}, question
            [{\begin{tabular}[t]{@{}l@{}} Instruction Following Ability\\ Benefits 
            Structure Understanding \S\ref{subsec:Instruction Following Ability Benefits Structure Understanding}\\ \end{tabular}} , middle
            ]
            [{\begin{tabular}[t]{@{}l@{}} Step-by-Step Reasoning Ability\\ Benefits 
            Structure Understanding \S\ref{subsec:Step-by-Step Reasoning Ability Benefits Schema Linking}\\ \end{tabular}} , middle
            ]
        ]
        [{\begin{tabular}[t]{@{}l@{}} How to Enhance\\Table Reasoning Ability\\in
        the Future \S\ref{sec:future}\end{tabular}}, question
            [Improving Table Reasoning Performance \S\ref{subsec:Improving Table Reasoning Performance}, future_middle
                [Supervised Fine-Tuning, technique_middle
                    [Establishing Diverse Training Data, future_leaf]
                ]
                [Result Ensemble, technique_middle
                    [Sampling Results More Efficiently, future_leaf]
                ]
                [In-context Learning, technique_middle
                    [Optimizing Prompts Automatically, future_leaf]
                ]
                [Instruction Design, technique_middle
                    [Automatically Refining Design with Verification, future_leaf]
                ]
                [Step-by-Step Reasoning, technique_middle
                    [Mitigating the Error Cascade in Multi-Step Reasoning, future_leaf]
                ]
            ]
            [Expanding Application \S\ref{subsec:Expanding Application}, future_middle
                [Multi-Modal, technique_middle
                    [Enhancing the Alignment between Image Tables and Questions, future_leaf]
                ]
                [Agent, technique_middle
                    [Cooperating with More Diverse and Suitable Table Agents, future_leaf]
                ]
                [Dialogue, technique_middle
                    [Backtracking the Sub-tables in the Multi-turn Interaction, future_leaf]
                ]
                [Retrieval-Augmented Generation, technique_middle
                    [Injecting Knowledge Related to the Entity, future_leaf]
                ]
            ]
        ]
    ]
\end{forest}
    }
    \caption{
        The structure overview of our paper, taking the most representative works as an example.
    }
    \label{fig:summary}
\end{figure*}

However, there is currently a lack of summary analysis on table reasoning works with LLMs, leading to how to improve the performance is still under exploration, which limits existing research to a certain extent.
Besides, the table reasoning survey of pre-LLMs is not suitable for LLMs. Since some mainstream techniques in the pre-LLM era, such as changing the model structure and designing pre-training tasks \cite{jin2022survey_tableqa}, are not suitable for LLM in table reasoning, while LLM methods focus more on designing prompts or pipelines \cite{LLMSurvey}. 
Therefore, this paper summarizes the existing works on table reasoning with LLMs to shed light on future research.
In detail, we focus on three questions of the table reasoning: \textit{1. What techniques can improve table reasoning performance in the LLM era}; \textit{2. Why LLMs excel at table reasoning}; \textit{3. How to enhance table reasoning ability in the future}.
The structure of this survey is shown in Figure~\ref{fig:summary}.

\textbf{Regarding the first topic}, to better adapt to table reasoning research in the LLM era, we introduce the mainstream techniques and detailed methods of table reasoning in the LLM era in $\S$\ref{sec:methodology}.
Specifically, we categorize existing works according to the different techniques that they utilize and detail them respectively.
\textbf{Considering the second topic}, we explore why LLMs show superior performance on table reasoning tasks in \S\ref{sec:discussion}.
We compare the best performance of pre-LLM and LLM on different benchmarks and prove that LLMs consistently surpass pre-LLMs in the table reasoning task.
Then, we discuss the advantages of LLMs in solving the table reasoning task based on the two inherent challenges of the task.
\textbf{Regarding the third topic}, we discuss the potential future directions of table reasoning in $\S$\ref{sec:future}.
To promote table reasoning research and better apply table reasoning to real-life scenarios, we separately analyze how to further improve table reasoning performance and explore how to adapt table reasoning into practical applications.

    \section{Background}
        \label{sec:background}
        \subsection{Paper Selection Criteria}
    To ensure the selected papers are highly related to the survey, the papers should meet the following criteria: 
    \textit{1.} Each question in the task that the paper aims to solve must be related to at least one table. \textit{2.} The method proposed in the paper is required to reason with or fine-tune LLMs.

\subsection{Task Definition}
    As the basis for subsequent analysis, in this section, we present the definition of the table reasoning task.
    In the table reasoning task, the input consists of the table, an optional text description, and the question tailored to the user requirement for various tasks (e.g., table QA, table fact verification, table-to-text, and text-to-SQL), and the output is the answer.
    
\subsection{Benchmarks}
    To help researchers understand the existing application scenarios of table reasoning in detail, we introduce four mainstream table reasoning tasks to which more than 90\% of the selected papers adapt, including text generation, entailment, and semantic parsing.
    An illustration of four tasks is shown in Figure~\ref{fig:intro}.
    Although most works of solving table reasoning tasks with LLMs do not need fine-tuning data, they still need to rely on labeled data to validate the performance.
    Therefore, in this subsection, we also provide one most-used validation benchmark for each task as an example and summarize the related resources in \url{https://github.com/zhxlia/Awesome-TableReasoning-LLM-Survey}:

    \begin{itemize}[nolistsep,leftmargin=*]
        \item \textbf{Table QA}: The table QA task is to answer a question according to a table \cite{pasupat-liang-2015-compositional_wikitablequestions}. 
        WikiTableQuestions~\cite{pasupat-liang-2015-compositional_wikitablequestions} serves as the initial benchmark in the table QA task, which has open-domain tables accompanied by complex questions.
        \item \textbf{Table Fact Verification}: The table fact verification task aims to verify whether a textual hypothesis is entailed or refuted based on the evidence tables \cite{2019TabFactA}. 
        TabFact~\cite{2019TabFactA}, as the first benchmark in the table fact verification task, features large-scale cross-domain table data and complex reasoning requirements.
        \item \textbf{Table-to-Text}: The table-to-text task is to generate a natural language description corresponding to the given question with a table \cite{nan-etal-2022-fetaqa}. 
        Different from the table QA task that only generates several spans, table-to-text requires the answer to be one paragraph.
        FeTaQA~\cite{nan-etal-2022-fetaqa} requires the model to generate a free-form answer to the question, with large-scale and high-quality data.
        \item \textbf{Text-to-SQL}: Text-to-SQL aims to convert a textual question under a database to executable structured query language (SQL). 
        Spider~\cite{yu-etal-2018-spider} is the first multi-domain, multi-table benchmark on the text-to-SQL task.
    \end{itemize} 

    \section{What Techniques Can Improve Table Reasoning Performance in the LLM Era}
        \label{sec:methodology}
        \begin{figure*}
    \centering
    \includegraphics[width=1\linewidth]{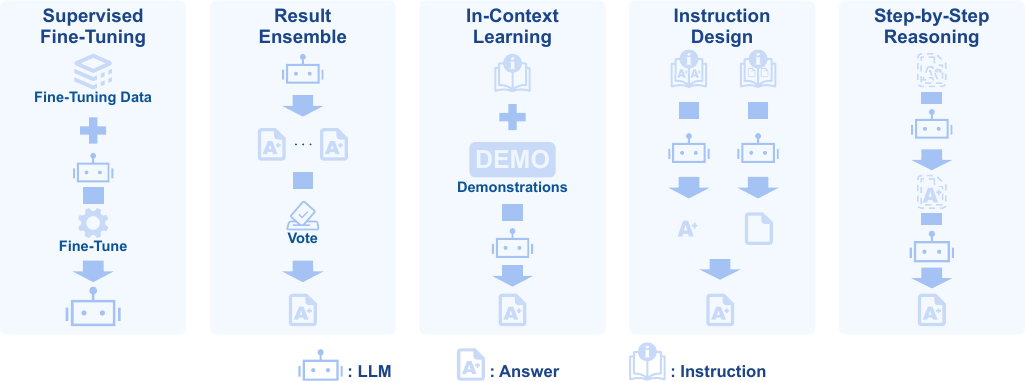}
    \caption{The mainstream techniques that can be utilized to improve table reasoning performance in the LLM era.}
    \label{fig:category}
\end{figure*}

There are significant differences between the model ability of the pre-LLM era and the LLM era, leading to the change in the mainstream techniques \cite{LLMSurvey}.
To help research better transition from the pre-LLM era to the LLM era, in this section, we discuss the mainstream techniques in the LLM era from two aspects: \textit{1}. the techniques following the pre-LLM era (\S\ref{subsec:Mainstream Techniques Following pre-LLMs}) and \textit{2}. the techniques unique to the LLM era (\S\ref{subsec:Mainstream Techniques Unique to LLMs}).
We categorize the table reasoning methods based on techniques they use into five categories, which are shown in Figure~\ref{fig:category}.
Then, we introduce the methods and highlight the changes in the technique, aiming to understand how to utilize the mainstream techniques in the LLM era.

\subsection{Mainstream Techniques Following pre-LLMs}
    \label{subsec:Mainstream Techniques Following pre-LLMs}
    Despite the considerable change in research brought about by LLMs, many pre-LLM techniques can still be applied to LLM.
    Therefore, we introduce the mainstream techniques following the pre-LLM era in this subsection.

    \subsubsection{Supervised Fine-Tuning}
        \label{subsubsec:supervised fine-tuning}
        Supervised fine-tuning refers to fine-tuning the LLM with annotated data to enhance the table reasoning capability.
        Since some open-source small-scale LLMs are weak in solving table tasks \cite{zhang2023tablellama} and have a relatively low cost of fine-tuning, researchers utilize the supervised fine-tuning techniques to enhance their performance.

        Existing works on supervised fine-tuning of LLM table reasoning include two types: 
        \textit{1}. leveraging \textbf{pre-existing or manually labeled data}, and 
        \textit{2}. leveraging \textbf{distilled data} generated by LLMs.
        Focusing on \textbf{pre-existing or manually labeled data}, to better complete the table reasoning task, TableGPT~\cite{zha2023tablegpt} fine-tunes the LLM by constructing instruction datasets.
        Considering the lack of generalization of previous work, TableLlama~\cite{zhang2023tablellama} constructs the training data by selecting representative table task datasets.
        Noting that annotating SQL data is too challenging, APEL~\cite{zhong-etal-2023-apel} proposes a method to annotate SQL, which generates the database according to schema and judges SQL correctness based on execution results.        
    
        Focusing on \textbf{distilled data}, \cite{yang2023distill} observes that the performance of the open-source model lags behind that of LLM on table-to-text tasks, thereby, this work utilizes the LLM as a teacher model to distill rationales and table descriptions and fine-tunes the open-source model with the distilled data.
        Besides, HELLaMA~\cite{bian2023hellama} concerns that some models could not locate the evidence based on the inputs, therefore it obtains training data to predict where the labeled description would be located by using other LLMs, and then fine-tune models.

        Based on pre-existing or manually labeled data, and distilled data embody the two thoughts to obtain training data in the LLM era.
        Pre-existing datasets are generally of high quality, but more limited in certain domains and tasks; whereas distilled data is less restrictive data but faces the problem of low-quality data.
        Therefore, how to significantly enhance the data quality of model distill through as little manual intervention as possible is an urgent issue to be studied.
        
        \paragraph{Highlight}
            The supervised fine-tuning methods in the pre-LLM era, limited by the model capabilities, can not bring about the generalization on unseen tasks \cite{xie-etal-2022-unifiedskg}.
            In contrast, for the LLM era, researchers design the instruction-based and multi-task data to fine-tune the model to enhance the table reasoning ability to generalize to different tasks, even tasks that are not seen in the training phase.
    
    \subsubsection{Result Ensemble}
        \label{subsubsec:robustness improvement}
    
        Result ensemble denotes improving table reasoning ability by selecting the most suitable answer from multiple results generated by LLM.
        Since the models of both the pre-LLM era and the LLM era could be less capable of maintaining correct results facing slight disturbance (e.g., random number seeds, meaningless words in questions), leading to model performance degradation \cite{ni2023lever}, researchers utilize the technique of result ensemble following the pre-LLM era.

        Existing methods of result ensemble in the LLM era mainly focus on two problems: 
        \textit{1}. how to \textbf{obtain diverse results} for a question, and 
        \textit{2}. how to \textbf{select the correct result} among the multiple results.
        Considering the work of \textbf{obtaining diverse results}, SQLPrompt~\cite{sun-etal-2023-sqlprompt} notes that the low diversity of results with fixed prompt and model causes results could be focused on specific incorrect answers, so proposes to generate results with multiple prompts and models.
        
        Regarding the work of \textbf{selecting the correct result}, Lever~\cite{ni2023lever} specifically trains a verifier to score each generated answer and selects the result with the highest score as the answer. 
        To select the correct from multiple candidate SQL queries, \cite{li2024using_rerank} proposes to construct test cases by generating new databases and using LLM to predict the execution results so that the test cases can distinguish all SQL with different execution results.
    
        These methods of solving the two problems can enhance the ensemble performance independently.
        Therefore, these two problems can be focused on together to further improve the table reasoning performance of the LLM.

        \paragraph{Highlight}

            Compared with pre-LLM methods, LLMs can generate more diverse results with more and simpler ways.
            For example, LLMs can obtain diverse results by only changing the instruction without changing the question, while pre-LLM methods have to ensure that the instructions of the fine-tuning and the inference are aligned \cite{gan-etal-2021-robustness_Text-to-SQL}.
            
\subsection{Mainstream Techniques Unique to LLMs}
    \label{subsec:Mainstream Techniques Unique to LLMs}
    In the LLM era, in addition to the mainstream techniques following the pre-LLM era, there are also techniques unique to LLM due to the emergence phenomenon \cite{LLMSurvey}.
    We present three typical emergent abilities mentioned in the previous research~\cite{LLMSurvey}.

    \subsubsection{In-context Learning}
        \label{subsubsec:in-context learning}
        The in-context learning refers to making the model generate the expected answer by using more suitable natural language instruction and multiple demonstrations (that is, the prompt), without requiring additional training or gradient updates \cite{LLMSurvey}.
        Since LLM performance is significantly affected by the prompt, researchers utilize the in-context learning technique by designing prompts to solve the table reasoning task directly.

        Regarding the work utilizing in-context learning ability in the table reasoning task, \cite{chen-2023-few} is the first to explore and demonstrate that LLM can reason about tables with in-context learning.
        ODIS~\cite{chang-fosler-lussier-2023-selective_odis} observes that in-domain demonstrations can improve model performance, so it synthesizes in-domain SQL based on SQL similarity.
        To address the challenge of demonstration selection, DAIL-SQL~\cite{gao2023dail}, and \cite{nan-etal-2023-enhancing} select demonstrations based on masked question similarity and SQL similarity respectively.
        To better parse complex tables, \cite{zhao-etal-2023-parseres} proposes to decode the table cells as a tuple to input that contains rich information.
        TAP4LLM~\cite{sui2023tap4llm} notices tables could contain noise and ambiguous information, therefore, it decomposes the table and then augments the sub-tables.
        Auto-CoT~\cite{zhang2023actsql_autocot} finds that the existing rationale annotation methods consume intensive resources, so uses the rule-based method of schema linking to generate rationales.

        \paragraph{Highlight}
            Because the models in the pre-LLM era can only learn fixed types of prompts through fine-tuning, it is hard to flexibly adjust prompts to enhance the reasoning performance of the model \cite{xie-etal-2022-unifiedskg}.
            Due to the in-context learning capability, LLMs can use various prompts that are suitable for different questions without further fine-tuning, which greatly reduces labeling overhead while enhancing performance.

    \subsubsection{Instruction Design}
        \label{subsubsec:instruction following}
    
        The instruction design denotes utilizing LLMs to solve tasks that are unseen during the training phase by designing the instruction description due to the instruction following ability of LLMs \cite{LLMSurvey}.
        In the table reasoning task, researchers utilize the instruction design technique to solve the task indirectly by instructing the LLM to complete multiple decomposed sub-tasks which could be novel and require the model to learn through instructions.      
        Existing works using the instruction design on table reasoning with LLMs focus on two types of methods: 
        \textit{1}. based on \textbf{modular decomposition}, and 
        \textit{2}. based on \textbf{tool using}.

        The researchers find that it is easier to complete decomposed sub-tasks than to complete the whole table reasoning task \cite{din-sql}, and LLM can generalize to different sub-tasks using the instruction following technique, thereby improving the performance of LLM on the table reasoning task by taking the method of modular decomposition.
        Both DATER~\cite{dater} and DIN-SQL~\cite{din-sql} note that decomposing table reasoning can effectively facilitate multi-step inference, thus they design pipelines for the table reasoning task to reduce the difficulty of inference.
        TableQAKit~\cite{lei2023tableqakit} identifies that Table QA tasks face different data and task forms, hindering the ease of research, so divide the Table QA task into a configuration module, a data module, a model module, and an evaluation module.
        In the open-domain setting, CRUSH4SQL~\cite{kothyari-etal-2023-crush4sql}, OpenTab~\cite{anonymous2023opentab}, and DB-GPT~\cite{xue2024dbgpt} decompose the task into two distinct phases, which are retrieving and reasoning to alleviate the problem of increased difficulty caused by extraneous irrelevant information.
        DBCopilot~\cite{wang2023dbcopilot} notices retrieval could suffer from the diverse expressions and vocabulary mismatch, so the task is decomposed into, firstly generating the question-relevant schema instead of retrieving and then reasoning.
        MAC-SQL~\cite{wang2023macsql} finds that the limited context window, single-pass generation, and the lack of verification result in poor performance, so the task is modularly decomposed into three modules to solve the problems.

        Faced with the decomposed sub-tasks of table reasoning, LLM, despite maintaining acceptable performance on most sub-tasks, does not excel at solving all sub-tasks (e.g., retrieval, numerical reasoning) \cite{cao-etal-2023-api}, so researchers instruct the LLM to invoke diverse tools to solve some sub-tasks, which is the method of tool using.
        StructGPT~\cite{jiang-etal-2023-structgpt} observes that the amount of structured data is too large to input to the model, so it provides different interfaces to extract multiple types of data, and the model obtains valid data by calling the appropriate interfaces.
        \cite{nan2023db_agent}, to explore and evaluate the action and reasoning capacities of LLM, proposes the long-form database question answering task, where LLMs need to decide an interaction strategy by reasoning, and then generate interaction commands to invoke the external model.
        To extend the model capabilities of various TableQA tasks, \cite{cao-etal-2023-api} enables querying knowledge and performing extra tabular operations by calling other LLM APIs.
        Also, some works focus on making tools and then employing them. Binder~\cite{binder}, noting that existing neural-symbolic works are model- and language-specific and require large training data, proposes to utilize the LLM to parse the sub-questions that are not translatable into the target program, such as SQL, then invoke the LLM to solve the sub-question.
        Recognizing the challenge of automatically transforming an arbitrary table in response to the question, ReAcTable~\cite{zhang2023reactable} proposes 
        leverage the LLM to generate a sequence of functions, which are then executed to produce an intermediate table, ultimately getting the answer.

        In summary, the methods of modular decomposition and tool using can be used together.
        Specifically, during solving the task with multiple modules, each module can enhance performance by employing tools.
        For example, about the retrieval modular, we can use programs to filter out the rows not related to the user question.
        
        \paragraph{Highlight}
            The pre-LLMs do not have the instruction following capability due to their weak generalization, where researchers have to train separate models for each sub-task when using the method of modular decomposition to solve table reasoning tasks \cite{unisar}.
            Also, it is hard to flexibly use or make diverse tools in the pre-LLM era \cite{LLMSurvey}.
            In contrast, LLM can achieve superior performance without individually fine-tuning for each sub-task or tool, saving the training overhead.

    \subsubsection{Step-by-Step Reasoning}
        The step-by-step reasoning indicates that solving complex reasoning tasks by employing prompt mechanisms that incorporate intermediate reasoning stages, referring to the technique and capability at the same time \cite{LLMSurvey}.
        The step-by-step reasoning, which requires the LLM to decompose the complex question into multiple simpler sub-questions, is different from modular decomposition, in which researchers need to break down tasks into widely different sub-tasks.
        MURMUR~\cite{saha-etal-2023-murmur} notices that prompting the LLM to reason step-by-step lacks explicit conditions between reasoning steps, proposes to first select the potentially correct models at each step, and then select the best model based on the score model.
        Chain-of-Table~\cite{anonymous2023chainoftable}, to reduce the difficulty of the single-hop reasoning, provides predefined table operations, from which LLMs select one operation and execute in each step.

        \paragraph{Highlight}
            %
            The methods of the pre-LLM era do not have the capability for step-by-step reasoning, so it is difficult to improve the performance of solving complex table reasoning by leveraging step-by-step reasoning.
            In contrast, LLM can decompose the reasoning into multiple steps, where the hardness of each step is lower than the full question, thereby decreasing the complexity of the table reasoning.

\subsection{Comparison}
    \subsubsection{Comparison of Technique Proportion}
        \begin{figure}
            \centering
            \resizebox{0.9\linewidth}{!}{
    \begin{tikzpicture} 
        \small
        \begin{axis}[
            xlabel=Month,
            ylabel=\#Paper,
            xmin=0, xmax=16,
            ymin=0, ymax=15,
            xtick={0,1,2,3,4,5,6,7,8,9,10,11,12,13,14,15,16},
            xticklabels={,22.10,,,23.01,,,23.04,,,23.07,,,23.10,,,24.01},
            legend style={
                at={(0.05,0.95)}, 
                anchor=north west, 
                font=\small, 
                legend cell align=left
            }
            ]
    
        \addplot[cgreen, line width = 0.5] plot coordinates {
            (0,0)
            (1,0)
            (2,0)
            (3,0)
            (4,0)
            (5,0)
            (6,0)
            (7,0)
            (8,0)
            (9,0)
            (10,1)
            (11,1)
            (12,2)
            (13,3)
            (14,5)
            (15,5)
            (16,5)
        };
        \addlegendentry{Supervised Fine-Tuning}
    
        \addplot[cblue, line width = 0.5] plot coordinates {
            (0,0)
            (1,0)
            (2,0)
            (3,0)
            (4,0)
            (5,1)
            (6,1)
            (7,1)
            (8,1)
            (9,1)
            (10,1)
            (11,1)
            (12,1)
            (13,1)
            (14,2)
            (15,2)
            (16,3)
        };
        \addlegendentry{Result Ensemble}
    
        \addplot[corange, line width = 0.5] plot coordinates {
            (0,0)
            (1,1)
            (2,1)
            (3,1)
            (4,1)
            (5,1)
            (6,1)
            (7,1)
            (8,2)
            (9,2)
            (10,2)
            (11,3)
            (12,3)
            (13,5)
            (14,5)
            (15,7)
            (16,7)
        };
        \addlegendentry{In-Context Learning}
    
        \addplot[cred, line width = 0.5] plot coordinates {
            (0,0)
            (1,1)
            (2,1)
            (3,1)
            (4,2)
            (5,2)
            (6,2)
            (7,3)
            (8,4)
            (9,4)
            (10,4)
            (11,4)
            (12,4)
            (13,7)
            (14,9)
            (15,12)
            (16,13)
        };
        \addlegendentry{Instruction Design}
    
        \addplot[cpurple, line width = 0.5] plot coordinates {
            (0,0)
            (1,0)
            (2,0)
            (3,1)
            (4,1)
            (5,1)
            (6,1)
            (7,1)
            (8,1)
            (9,1)
            (10,1)
            (11,1)
            (12,1)
            (13,1)
            (14,1)
            (15,1)
            (16,2)
        };
        \addlegendentry{Step-by-Step Reasoning}
        
        \end{axis}
    \end{tikzpicture}
}
            \caption{
                The research trend using different techniques over months.
                \#Paper denotes the number of papers.
            }
            \label{fig:method_distribution}
        \end{figure}
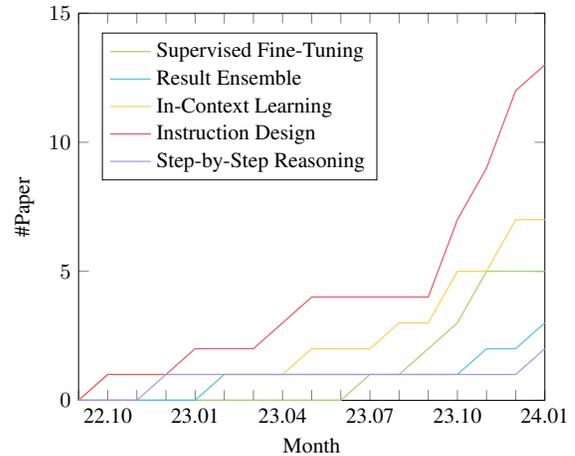
        To analyze the research trend of existing studies on table reasoning with LLMs, we static the paper number of existing studies depending on the used technique as we know, which is shown in Figure~\ref{fig:method_distribution}.
        From the figure, it can be found that \textit{studying the instruction design and in-context learning is more promising in the table reasoning task than studying step-by-step reasoning and result ensemble}.
        That is because the works on step-by-step reasoning and result ensemble account for a relatively small proportion because these types of work can easily be applied to different tasks, fewer researchers focus solely on table reasoning tasks using the result ensemble technique, which is discussed in detail in \S\ref{sec:future}.     
        On the contrary, the instruction design and in-context learning techniques need methods to be designed specifically for the table reasoning task and have a lower time overhead than using the supervised fine-tuning technique, so the works of the instruction design and in-context learning are the most common among table reasoning studies.

    \subsubsection{Comparison of Technique Performance}
        \begin{table*}[t]
            \centering
            \small
            \resizebox{\textwidth}{!}{
                \begin{tabular}{l|llll}
                    \toprule
                    \textbf{Mainstream Techniques} & \textbf{WikiTQ$^\dag$} & \textbf{TabFact} & \textbf{FeTaQA} & \textbf{Spider} \\
                    \midrule
                    Supervised Fine-Tuning & $0.32$~\cite{zhang2023tablellama} & $0.83$~\cite{zhang2023tablellama} & $0.67$~\cite{bian2023hellama} & - \\
                    Result Ensemble & $0.66$~\cite{ni2023lever} & - & - & - \\
                    In-Context Learning & - & $0.60$~\cite{sui2023tap4llm} & - & \bm{$0.87$}~\cite{gao2023dail} \\
                    Instruction Design & \bm{$0.68$}~\cite{zhang2023reactable} & \bm{$0.93$}~\cite{dater} & \bm{$0.71$}~\cite{zhang2023reactable} & $0.85$~\cite{din-sql} \\
                    Step-by-Step Reasoning & $0.67$~\cite{anonymous2023chainoftable} & $0.87$~\cite{anonymous2023chainoftable} & $0.66$~\cite{anonymous2023chainoftable} & - \\
                    \bottomrule
                \end{tabular}
            }
            \caption{The best results on different benchmarks under each mainstream technique. $^\dag$ refers to the WikiTableQuestions. The evaluation metric for WikiTableQuestions/TabFact/FeTaQA/Spider is accuracy/accuracy/ROUGE-1/execution accuracy.}
            \label{tab:category_sota}
        \end{table*}
    
        To analyze the most effective techniques, thereby finding the promising research directions, we static the highest scores achieved by LLM methods using different mainstream techniques on different benchmarks in Table~\ref{tab:category_sota}.
        It can be found that instruction design and step-by-step reasoning improve the table reasoning capability of LLMs across different tasks consistently, which we discuss in detail in \S\ref{sec:discussion}.
        In addition, the consistency of performance improvement across different tasks also shows that the ability required between different table reasoning tasks has a high consistency, requiring the high generalization of LLMs.
        It is worth noting that in-context learning achieves the best performance in the text-to-SQL task because SQL has a simpler syntax compared with the natural language, wherewith the same number of demonstrations, the text-to-SQL task can cover more types of user questions, attracting more attention than solving other table reasoning tasks with in-context learning.

    \section{Why LLMs Excel at Table Reasoning}
        \label{sec:discussion}
        \begin{table}[t]
    \centering
    \small
    \resizebox{\linewidth}{!}{
        \begin{tabular}{l|ll}
            \toprule
            \textbf{Benchmarks} & \textbf{pre-LLM} & \textbf{LLM} \\
            \midrule
            WikiTQ$^\dag$ & $0.63$~\cite{jiang-etal-2022-omnitab} & \bm{$0.68$}~\cite{zhang2023reactable} \\
            TabFact & $0.85$~\cite{zhao-yang-2022-tablefact_plm_lka} & \bm{$0.93$}~\cite{dater} \\
            FeTaQA & $0.65$~\cite{xie-etal-2022-unifiedskg} & \bm{$0.71$}~\cite{zhang2023reactable} \\
            Spider & $0.80$~\cite{resdsql} & \bm{$0.87$}~\cite{gao2023dail} \\  
            \bottomrule
        \end{tabular}
    }
    \caption{The best performance of pre-LLM and LLM methods on different benchmarks. $^\dag$ denotes WikiTableQuestions. The evaluation metric for WikiTableQuestions/TabFact/FeTaQA/Spider is accuracy/accuracy/ROUGE-1/execution accuracy.}
    \vspace{-1em}
    \label{tab:sota}
\end{table}

LLMs surpass pre-LLMs (Table~\ref{tab:sota}) in table reasoning by the methods in \S\ref{sec:methodology}. 
We analyze the key insights behind this from structure understanding and schema linking, which are two main challenges of the table reasoning \cite{tabert}.

\subsection{Instruction Following Ability Benefits Structure Understanding}
    \label{subsec:Instruction Following Ability Benefits Structure Understanding}
    Structure understanding means understanding the table schema (e.g., columns, rows) and their relationships, which provides the key evidence and necessary context information for decoding \cite{tabert}.
    Compared with pre-LLMs, LLMs can solve the challenge of structure understanding better, mainly due to the instruction following ability.
    For example, the code parsing ability brought by the instruction following could benefit table understanding ability because both require recognizing the hierarchical structure from plain input (e.g., linearized table to structured table, contextualized code to structured code) \cite{cao-etal-2023-api}.

    
    
\subsection{Step-by-Step Reasoning Ability Benefits Schema Linking}
    \label{subsec:Step-by-Step Reasoning Ability Benefits Schema Linking}
    Schema Linking refers to aligning the entity mentioned in question with the entity in tables \cite{tabert}.
    Compared with pre-LLM, LLMs have stronger capabilities of schema linking, mainly for the step-by-step reasoning ability of the LLM.
    Specifically, LLMs can simplify the linking from sentence-level to span-level, via decomposing the complete question and table and filtering the irrelevant context \cite{din-sql}.

    
    \section{How to Enhance Table Reasoning Ability in the Future}
        \label{sec:future}
To promote table reasoning research in the LLM era and apply table reasoning to actual scenarios, we discuss the future research directions in this section from both enhancing table reasoning and expanding practical applications.

\subsection{Improving Table Reasoning Performance}
    \label{subsec:Improving Table Reasoning Performance}
    Although the existing LLM-based method has significantly improved performance compared with the pre-LLM era, there is still a certain gap in the thorough solution of the table reasoning task.
    Therefore, in this subsection, we analyze the shortcomings and possible improvements of existing works on the table reasoning task under each category in \S\ref{sec:methodology}.

    \paragraph{Supervised Fine-Tuning: Establishing Diverse Training Data}
        Due to the strong generalization of LLMs, researchers should construct diverse data for multiple table tasks when performing supervised fine-tuning of LLMs to improve the overall performance on table reasoning tasks.
        As the discussion in \S\ref{subsubsec:supervised fine-tuning}, current pre-existing or manually labeled data methods simply mix diverse data from different table tasks as training data to fine-tune the LLMs.
        However, the proportion of different tasks in the training data has a significant impact on model performance.
        Future work should balance the diverse training data from multiple tasks in different proportions to explore the optimal proportion for optimizing the table reasoning capabilities of fine-tuning LLMs.
    
        Apart from labeling data, existing methods of distilled data only focus on certain features or specific tasks, resulting in a lack of diversity in the distilled data, and the table reasoning performance of the model cannot be comprehensively improved by fine-tuning with the distilled data.
        Therefore, it is worth exploring how to distill diverse data for different tasks to improve the LLM comprehensive ability and generalization in table reasoning tasks.
    
    \paragraph{Result Ensemble: Sampling Results More Efficiently}
        To obtain the correct answer after ensemble, researchers should focus on how to sample in the possible result space effectively.
        The main purpose of obtaining multiple results is to widen the sampling space so that the correct answer can be sampled multiple times.
        However, existing works do not consider changing the demonstrations in the prompt to improve the correctness of the results, and the impact of the demonstrations on the table reasoning performance of LLMs is significant.
        Future work should change the demonstrations to sample results that are more likely to be correct.
        
        Current studies on selecting the correct answer only rely on the final result and do not take into account that the number of results increases exponentially with the growing number of reasoning steps, and it is difficult to sample the correct answer in an exponentially large search space.
        Future work should narrow the search space by selecting the correct reasoning path at each step, and then selecting the correct answer based on the searched path \cite{xie2023selfevaluation}.

    \paragraph{In-Context Learning: Optimizing Prompts Automatically}
        Since the in-context learning performance of LLMs relies heavily on prompts, researchers should focus on how to automatically optimize prompts for table reasoning based on the question.
        Prompt design research on single-step reasoning compares candidate prompts from a limited range of human-labeled instructions and examples, which results in performance improvement is also limited.
        To design a better prompt, future work should automatically generate and optimize the prompt based on the questions and tables.
    
    \paragraph{Instruction Design: Automatically Refining Design with Verification}
        Depending on the discussion in \S\ref{subsubsec:instruction following}, how to make fuller use of the capability of the instruction following to reduce the difficulty of each table reasoning question deserves the attention of researchers.
        Current methods of modular decomposition require manually decomposing the task into different modules in advance.
        However, this decomposition can only apply to a certain table task. 
        In contrast, the fixed decomposition applicable to all table tasks is too general and does not reduce the difficulty of reasoning well.
        Therefore, rather than specifying the decomposition for a particular table task, future work should automatically decompose the task according to the question, which is suitable for all table tasks without human involvement and greatly reduces the difficulty degree of single-step reasoning.

        For the methods of tool using, current works do not notice that the process of invoking tools may cause extra errors in the table reasoning process.
        Future work should include a tool verification process that prompts the LLMs to revise the tools to ensure that the tools can be applied correctly in the table reasoning task, thereby enhancing the accuracy.
        
    \paragraph{Step-by-Step Reasoning: Mitigating the Error Cascade in Multi-Step Reasoning}
        Existing studies on step-by-step reasoning do not consider the error cascade problem in table reasoning and cause erroneous intermediate results leading to errors in subsequent reasoning.
        The prompt method of Tree-of-Thought~\cite{yao2023treeofthoughts} (ToT) alleviates this problem by maintaining multiple possible intermediate steps in multi-step reasoning, so how to apply ToT to table reasoning tasks deserves future attention.

\subsection{Expanding Application}
    \label{subsec:Expanding Application}
    In this subsection, we analyze the requirements of table reasoning tasks in real-life scenarios and propose future expandable directions accordingly.

    \paragraph{Multi-Modal: Enhancing the Alignment between Image Tables and Questions}
        The multi-modal setting requires the model to encompass automated comprehension, classification, and extraction of information from textual, visual, and other forms of evidence.
        Because there are many tables stored in the form of images in actual scenarios, direct Optical Character Recognition (OCR) will cause information loss due to the recognition errors, so we need to combine visual models to better understand and reason about image tables.
        To better align visual information and natural language questions, future research can explore design structures to align entities in questions with headers in image tables, thereby enhancing semantic alignment between images and text.
    
    \paragraph{Agent: Cooperating with More Diverse and Suitable Table Agents}
        The agent denotes an entity equipped with the capabilities to perceive the surrounding environment, engage in decision-making processes, and execute actions based on these decisions \cite{xi2023agent_survey}.
        In real scenarios, when LLM faces complex table reasoning problems that are difficult to solve alone, it can cooperate with other agents, such as code, and search engines.
        Because different agents are suitable for solving different tasks and bring different performance changes to the same task, future research can enhance cooperation with agents by exploring more diverse agents suitable for different table tasks in actual scenarios \cite{cao-etal-2023-api}.
    
    \paragraph{Dialogue: Backtracking the Sub-tables in the Multi-turn Interaction}
        Dialogue systems are designed to converse with humans as assistants through conversational interactions.
        When interacting with users, there could be problems such as incorrect model results and ambiguous questions, which require multiple turns to correct errors. 
        However, in the LLM era, few researchers pay attention to the table reasoning task of multi-turn dialogue.
        Therefore, it is necessary to explore table reasoning with dialogues.
        The model needs to focus on the sub-tables related to the user question, especially when facing huge tables \cite{dater}. 
        During multiple turns of dialogues, the question-related sub-tables are constantly changing, thereby future work should study how to backtrace the sub-tables decomposed to obtain the whole relevant information, preventing the last sub-table not including the required information in the turn \cite{yao2023treeofthoughts}.
    
    \paragraph{Retrieval-Augmented Generation: Injecting Knowledge Related to the Entity}
        Retrieval-Augmented Generation (RAG) technology denotes retrieving the reasoning-related information from a large number of documents before reasoning \cite{gao2024rag_survey}.
        Since the table reasoning task often faces knowledge-intensive scenarios in applications where the in-domain knowledge of LLM is not sufficient to solve, future work should focus on enhancing table reasoning capabilities by retrieving knowledge.
        In table reasoning tasks, LLMs could be challenging to understand the meaning of some entities in the table, thereby lowering the answer accuracy \cite{guo-etal-2019-schema_linking}. 
        To solve this challenge, future research should detect the unknown entities in the table and inject corresponding knowledge related to such entities.



    \section{Conclusion}
        In this paper, we summarize existing research work on table reasoning with LLMs.
        In the LLM era, the supervised fine-tuning and result ensemble methods following the pre-LLM era are still effective. 
        Besides, the in-context learning, instruction following, and step-by-step reasoning techniques unique to the LLM era can also be used to improve the model table reasoning performance.
        Also, LLMs surpass pre-LLMs in table reasoning tasks because of the instruction following and step-by-step reasoning capabilities of LLMs.
        To inspire future research, we explore potential future directions for improving table reasoning performance. 
        We also explore four future improvement directions for real applications.
        Finally, we summarize the current resources of the table reasoning in \href{https://github.com/zhxlia/Awesome-TableReasoning-LLM-Survey}{GitHub} and will continue to update it.

    \clearpage
    \bibliographystyle{named}
    \bibliography{ijcai24}
    
\end{document}